\documentclass[runningheads,a4paper]{llncs}
\usepackage{standalone}

\makeatletter
\def\@documentnocite#1{\@bsphack
  \@for\@citeb:=#1\do{%
    \edef\@citeb{\expandafter\@firstofone\@citeb}%
    \if@filesw\immediate\write\@auxout{\string\citation{\@citeb}}\fi
    \@ifundefined{b@\@citeb}{\G@refundefinedtrue
      \@latex@warning{Citation `\@citeb' undefined}}{}}%
  \@esphack}
\AtBeginDocument{\let\nocite\@documentnocite}
\makeatother

\usepackage[utf8]{inputenc}
\usepackage{verbatim}
\usepackage{multicol}
\usepackage{amsmath}
\usepackage{amsfonts}
\usepackage{amssymb}
\usepackage{graphicx}
\usepackage{lmodern}
\usepackage{calc}
\usepackage{enumitem}
\usepackage{algpseudocode}
\usepackage{algorithm}
\usepackage{algorithmicx}

\algsetblockdefx[IfContinue]{IfContinue}{IfContinue}
{0}{0pt}
[0]{}
[1]{\textbf{if} #1 \textbf{continue}}

\algrenewcommand\algorithmicrequire{%
  \makebox[\widthof{\textbf{Output:}}][l]{\textbf{Input:}}}
  
 \algrenewcommand\algorithmicensure{%
  \textbf{Output:}}

\usepackage{color}
\usepackage{gnuplottex}
\usepackage[font=small,labelfont=bf]{caption}
\usepackage[font=scriptsize,labelfont=bf]{subcaption}
\usepackage{microtype}
\usepackage[normalem]{ulem}
\usepackage{tikz}
\usetikzlibrary{trees,automata,positioning}
\usepackage{booktabs}
\usepackage{gnuplottex}
\usepackage{xparse}
\usepackage{epstopdf}
\ExplSyntaxOn
\DeclareExpandableDocumentCommand{\convertlen}{ O{cm} m }
 {
  \dim_to_unit:nn { #2 } { 1 #1 } cm
 }
\ExplSyntaxOff

\tikzstyle{level 1}=[level distance=3.0cm, sibling distance=0.6cm]
\tikzstyle{level 2}=[level distance=3.5cm, sibling distance=0.6cm]
\tikzstyle{level 3}=[level distance=3.5cm, sibling distance=0.6cm]

\tikzstyle{l1} = [rectangle, text width=5em, text centered]
\tikzstyle{l2} = [rectangle, text width=5em, text centered]
\tikzstyle{l3} = [rectangle, text width=5em, text centered]

\author{Micky Faas \and Matthijs van Leeuwen}
\title{Vouw: Geometric Pattern Mining\\using the MDL Principle}
\institute{Leiden Institute for Advanced Computer Science, Leiden University}

\begin{document}
\mainmatter
\maketitle

\begin{abstract}
We introduce geometric pattern mining, the problem of finding recurring local structure in discrete, geometric matrices. It differs from existing pattern mining problems by identifying complex spatial relations between elements, resulting in arbitrarily shaped patterns. After we formalise this new type of pattern mining, we propose an approach to selecting a set of patterns using the Minimum Description Length principle. We demonstrate the potential of our approach by introducing Vouw, a heuristic algorithm for mining exact geometric patterns. We show that Vouw delivers high-quality results with a synthetic benchmark.
\end{abstract}

\section{Introduction}

Frequent pattern mining \cite{aggarwal2014fpm} is the well-known subfield of data mining that aims to find and extract recurring substructures from data, as a form of knowledge discovery. The generic concept of pattern mining has been instantiated for many different types of patterns, e.g., for item sets (in Boolean transaction data), subgraphs (in graphs/networks), and episodes (in sequences). So far, however, little research has been done on pattern mining for raster-based data, i.e., geometric matrices in which the row and column orders are fixed. The exception is geometric tiling \cite{gionis2004tiles,tatti2012stijl}, but that problem only considers tiles, i.e., rectangular-shaped patterns, in Boolean data.

\begin{figure}[b]
\centering
\begin{subfigure}[t]{0.35\textwidth}
\centering
\includegraphics[scale=.25]{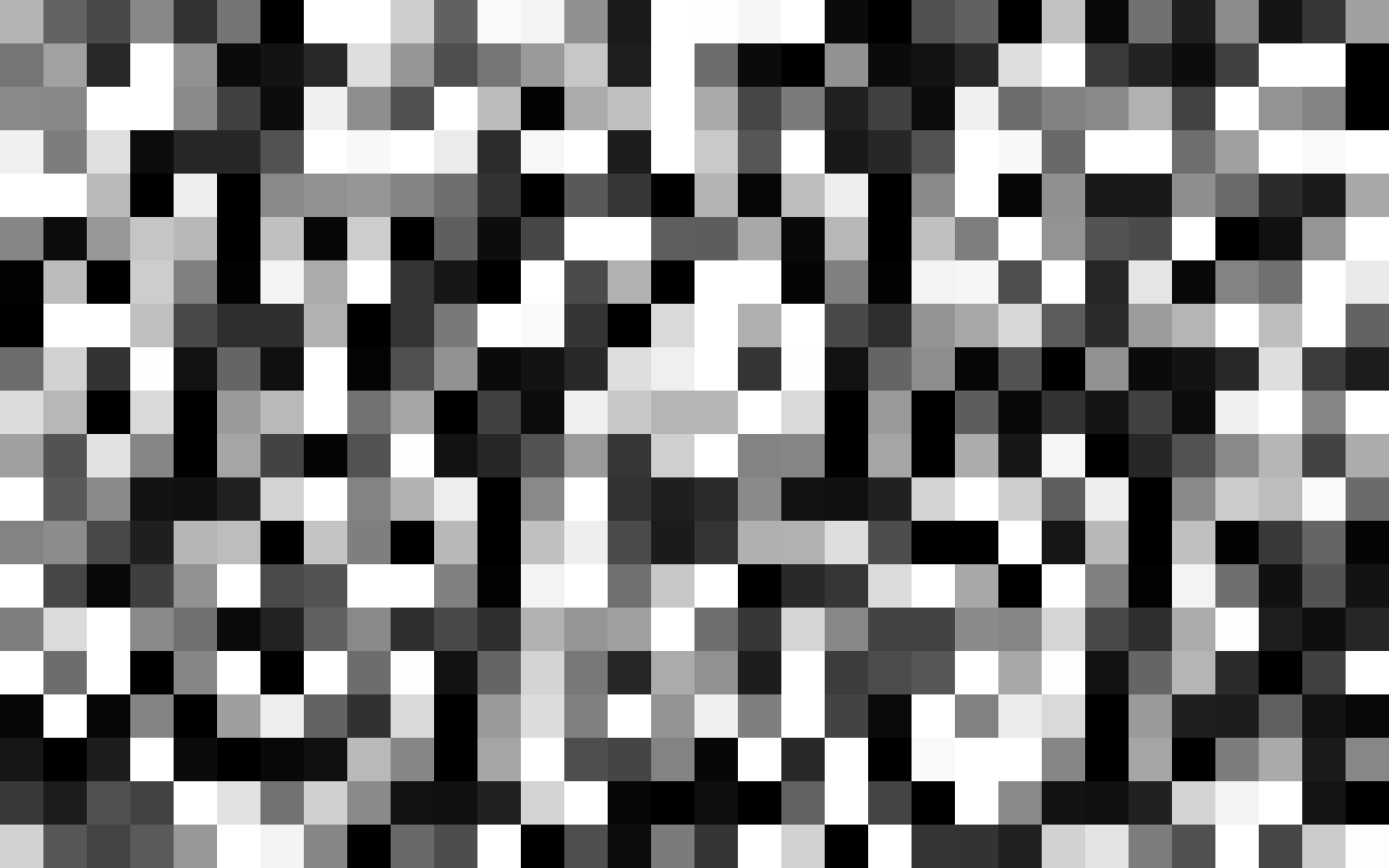}
\caption{$32\times 24$ `geometric matrix'.}
\label{fig-example1a}
\end{subfigure}%
~
\begin{subfigure}[t]{0.21\textwidth}
\centering
\includegraphics[scale=.25]{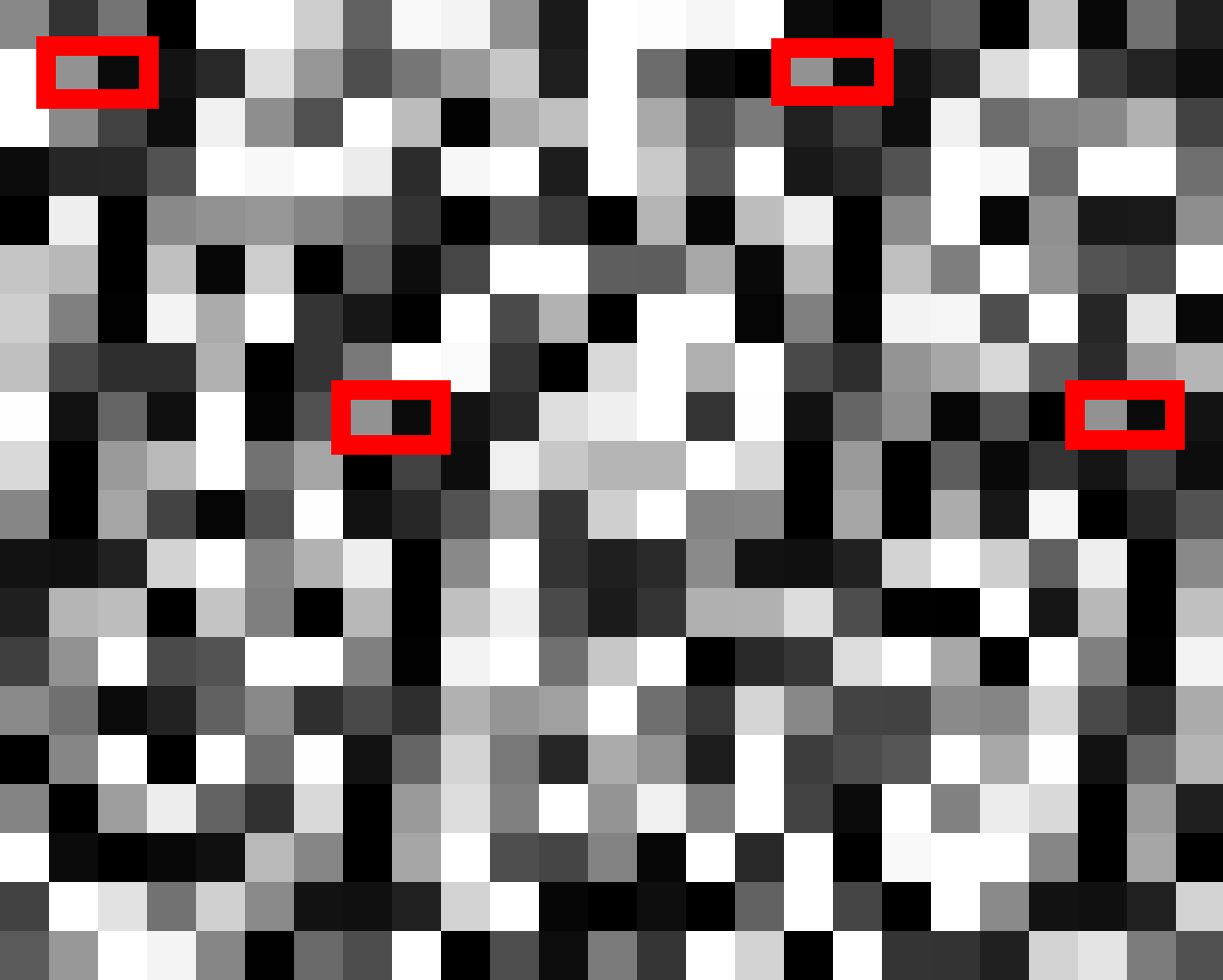}
\caption{Pair $(146,11)$.}
\label{fig-example1b}
\end{subfigure}%
~
\begin{subfigure}[t]{0.37\textwidth}
\centering
\includegraphics[scale=.25]{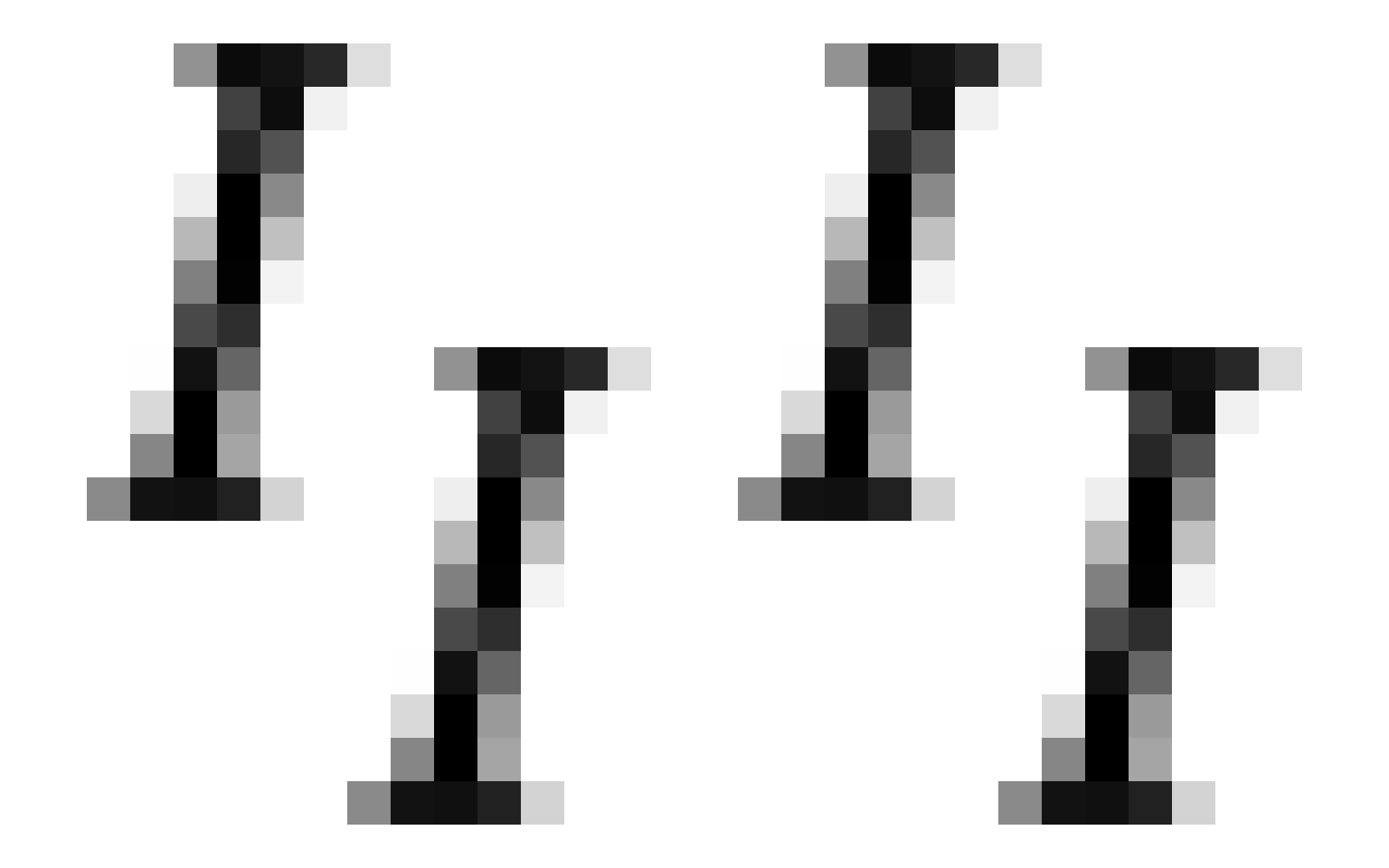}
\caption{Pattern `I' occurs four times.}
\label{fig-example1c}
\end{subfigure}%
\caption{Geometric pattern mining example. Each element is in $[0,255]$.}
\label{fig-example1}
\end{figure}  

In this paper we generalise this setting in two important ways. First, we consider geometric patterns \emph{of any shape} that are geometrically connected, i.e., it must be possible to reach any element from any other element in a pattern by only traversing elements in that pattern. Second, we consider \emph{discrete geometric data} with any number of possible values (which includes the Boolean case). We call the resulting problem \emph{geometric pattern mining}.

Figure~\ref{fig-example1} illustrates an example of geometric pattern mining.  Figure~\ref{fig-example1a} shows a $32 \times 24$ grayscale `geometric matrix', with each element in $[0,255]$, apparently filled with noise. If we take a closer look at all horizontal pairs of elements, however, we find that the pair $(146,11)$ is, amongst others, more prevalent than expected from `random noise' (Figure~\ref{fig-example1b}). If we would continue to try all combinations of elements that `stand out' from the background noise, we would eventually find four copies of the letter `I' set in 16 point Garamond Italic (Figure~\ref{fig-example1c}).

The 35 elements that make up a single `I' in the example form what we call a \emph{geometric pattern}. Since its four occurrences jointly cover a substantial part of the matrix, we could use this pattern to describe the matrix more succinctly than by 768 independent values. That is, we could describe it as the pattern `I' at locations $(5,4), (11,11), (20,3), (25,10)$ plus 628 independent values, hereby separating structure from accidental (noise) data. Since the latter description is shorter, we have compressed the data. At the same time we have learned something about the data, namely that it contains four I's. This suggests that we can use compression as a criterion to find patterns that describe the data.

\smallskip
\noindent \textbf{Approach and contributions}. Our first contribution is that we introduce and formally define \emph{geometric pattern mining}, i.e., the problem of finding recurring local structure in geometric, discrete matrices. Although we restrict the scope of this paper to two-dimensional data, the generic concept applies to higher dimensions. Potential applications include the analysis of satellite imagery, texture recognition, and (pattern-based) clustering.

We distinguish three types of geometric patterns: 1) \emph{exact} patterns, which must appear exactly identical in the data to match; 2) \emph{fault-tolerant} patterns, which may have noisy occurrences and are therefore better suited to noisy data; and 3) \emph{transformation-equivalent} patterns, which are identical after some transformation (such as mirror, inverse, rotate, etc.). Each consecutive type makes the problem more expressive and hence more complex. In this initial paper we therefore restrict the scope to the first, exact type.

As many geometric patterns can be found in a typical matrix, it is crucial to find a compact set of patterns that together describe the structure in the data well. We regard this as a model selection problem, where a model is defined by a set of patterns. Following our observation above, that geometric patterns can be used to compress the data, our second contribution is the formalisation of the model selection problem by using the \emph{Minimum Description Length (MDL) principle} \cite{rissanenmdl,grunwaldmdl}. Central to MDL is the notion that `learning' can be thought of as `finding regularity' and that regularity itself is a property of data that is exploited by \emph{compressing} said data. This matches very well with the goals of pattern mining, as a result of which the MDL principle has proven very successful for MDL-based pattern mining \cite{krimp,classy}.

Finally, our third contribution is Vouw, a heuristic algorithm for MDL-based geometric pattern mining that (1) finds compact yet descriptive sets of patterns, (2) requires no parameters, and (3) is tolerant to noise in the data (but not in the occurrences of the patterns). We empirically evaluate Vouw on synthetic data and demonstrate that it is able to accurately recover planted patterns.

\section{Related Work}

As the first pattern mining approach using the MDL principle, Krimp \cite{krimp} was one of the main sources of inspiration for this paper. Many papers on pattern-based modelling using MDL have appeared since, both improving search, e.g., Slim \cite{slim}, and extensions to other problems, e.g., Classy \cite{classy} for rule-based classification.

The problem closest to ours is probably that of geometric tiling, as introduced by Gionis et al.\ \cite{gionis2004tiles} and later also combined with the MDL principle by Tatti and Vreeken \cite{tatti2012stijl}. Geometric tiling, however, is limited to Boolean data and rectangularly shaped patterns (tiles); we strongly relax both these limitations (but as of yet do not support patterns based on densities or noisy occurrences).

Campana et al. \cite{campana2010compression} also use matrix-like input data (textures) and develop a compression-based similarity measure. Their method, however, cannot be used for \emph{explanatory} data analysis as it relies on a generic image compression algorithm that is essentially a black box.



\section{Geometric Pattern Mining using MDL}



We define geometric pattern mining on bounded, discrete and two-dimensional raster-based data. We represent this data as an $M\times N$ matrix $A$ whose rows and columns are finite and in a fixed ordering (i.e., reordering rows and columns semantically alters the matrix). Each element $a_{i,j} \in S$, where row $i \in [0;N)$, column $j \in [0;M)$, and $S$ is a finite set of symbols, i.e., the alphabet of $A$. 

\begin{figure}[b]
\input{"figures/decomposition_example.tex"}
\label{example1}
\end{figure}

According to the MDL principle, the shortest (optimal) description of $A$ reveals all structure of $A$ in the most succinct way possible. This optimal description is only optimal if we can unambiguously reconstruct $A$ from it and nothing more---the compression is both minimal and lossless. Figure~\ref{example1} illustrates how an example matrix could be succinctly described using patterns: matrix $A$ is decomposed into patterns $X$ and $Y$. A set of such patterns constitutes the \textbf{model} for a matrix $A$, denoted $H_A$ (or $H$ for short when $A$ is clear from the context). In order to reconstruct $A$ from this model, we also need a mapping from the $H_A$ back to $A$. This mapping represents what (two-part) MDL calls the \textbf{the data given the model $H_A$}. In this context we can think of this as a set of all instructions required to rebuild $A$ from $H_A$, which we call the \textbf{instantiation} of $H_A$ and is denoted by ${I}$ in the example. These concepts allow us to express matrix $A$ as a decomposition into sets of local and global spatial information, which we will next describe in more detail.



\subsection{Patterns and Instances}
\noindent $\triangleright$ \emph{We define a \textbf{pattern} as an $M_X\times N_X$ submatrix $X$ of the original matrix $A$. Elements of this submatrix may be $\cdot$, the empty element, which gives us the ability to cut-out any irregular-shaped part of $A$. We additionally require the elements of $X$ to be adjacent (horizontal, vertical or diagonal) to at least one non-empty element and that no rows and columns are empty.}

\smallskip

From this definition, the dimensions $M_X\times N_X$ give the smallest rectangle around $X$ (the \emph{bounding box)}. We also define the cardinality $|X|$ of $X$ as the number of non-empty elements. We call a pattern $X$ with $|X|=1$ a \textbf{singleton pattern}, i.e., a pattern containing exactly one element of $A$. 

Each pattern contains a special \textbf{pivot} element: %
$pivot(X)$ is the first non-empty element of $X$. 
\noindent
A pivot can be thought of as a fixed point in $X$ which we can use to position its elements in relation to $A$. This translation, or \textbf{offset}, is a tuple ${q}=(i,j)$ that is on the same domain as an index in $A$. We realise this translation by placing all elements of $X$ in an empty $M\times X$ size matrix such that the pivot element is at $(i,j)$. We formalise this in the \textbf{instantiation operator} $\otimes$:

\smallskip
\noindent $\triangleright$
\emph{We define the \textbf{instance} $X \otimes {(i,j)}$ as the $M\times N$ matrix containing all elements of $X$ such that $\mathrm{pivot}(X)$ is at index $(i,j)$ and the distances between all elements are preserved. The resulting matrix contains no additional non-empty elements. } %
\smallskip

Since this does not yield valid results for arbitrary offsets $(i,j)$, we enforce two constraints: (1) an instance must be \textbf{well-defined}: placing $\mathrm{pivot}(X)$ at index $(i,j)$ must result in an instance that contains all elements of $X$, and (2) elements of instances cannot \emph{overlap}, i.e., each element of $A$ can be described only once. 

\smallskip
\noindent $\triangleright$
\emph{Two pattern instances $X \otimes {q}$ and $Y \otimes {r}$, with ${q} \neq {r}$ are \textbf{non-overlapping} if $|(X \otimes {q}) + (Y \otimes {r})| = |X|+|Y|$.}
\smallskip

From here on we will use the same letter in lower case to denote an arbitrary instance of a pattern, e.g., $x = X \otimes {q}$ when the exact value of ${q}$ is unimportant. Since instances are simply patterns projected onto an $M\times N$ matrix, we can reverse $\otimes$ by removing all completely empty rows and columns:

\smallskip
\noindent $\triangleright$
\emph{Let $X \otimes {q}$ be an instance of $X$, then by definition we say that $\oslash(X \otimes {q}) = X$.}
\smallskip

We briefly introduced the instantiation $I$ as a set of `instructions' of where instances of each pattern should be positioned in order to obtain $A$. As Figure \ref{example1} suggests, this mapping has the shape of an $M\times N$ matrix.

\smallskip
\noindent $\triangleright$
\emph{Given a set of patterns $H$, the \textbf{instantiation (matrix)} ${I}$ is an $M\times N$ matrix such that ${I}_{i,j} \in H \cup \{\cdot\}$ for all $(i,j)$, where $\cdot$ denotes the empty element. For all non-empty ${I}_{i,j}$ it holds that ${I}_{i,j} \otimes (i,j)$ is a non-overlapping instance of ${I}_{i,j}$ in $A$.}
\subsection{The Problem and its Solution Space}

\label{constructpatterns}

Larger patterns can be naturally constructed by joining (or merging) smaller patterns in a bottom-up fashion. 
To limit the considered patterns to those relevant to $A$, instances can be used as an intermediate step. As Figure \ref{example2} demonstrates, we can use a simple element-wise matrix addition to sum two instances and use $\oslash$ to obtain a joined pattern. Here we start by instantiating $X$ and $Y$ with offsets $(1,0)$ and $(1,1)$, respectively. We add the resulting ${x}$ and ${y}$ to obtain $\oslash{z}$, the union of $X$ and $Y$ with relative offset $(1,1)-(1,0)=(0,1)$. 

\begin{figure}[t]
\centering
\input{"figures/pattern_construction.tex"}
\label{example2}
\end{figure}



\smallskip
\noindent \textbf{The Sets $\mathcal{H}_A$ and $\mathcal{I}_A$}.\label{thesetH}
We define the \textbf{model class} $\mathcal{H}$ as the set of all possible models for all possible inputs. Without any prior knowledge, this would be the search space. To simplify the search, however, we only consider the more bounded subset $\mathcal{H}_A$ of all possible models for $A$, and $\mathcal{I}_A$, the set of all possible instantiations for these models. To this end we first define $H_A^0$ to be the model with only singleton patterns, i.e., $H_A^0=S$, and denote its corresponding instantiation matrix by ${I}_A^0$. Given that each element of ${I}_A^0$ must correspond to exactly one element of $A$ in $H_A^0$, we see that each ${I}_{i,j} = a_{i,j}$ and so we have ${I}_A^0 = A$. 

Using $H_A^0$ and ${I}_A^0$ as base cases we can now inductively define $\mathcal{I}_A$: 
\vspace{-0.2\baselineskip}
\begin{description}[labelwidth=\widthof{\bfseries By induction}]
\item[Base case] ${I}_A^0 \in \mathcal{I}_A$
\item[By induction] If ${I}$ is in $\mathcal{I}_A$ then take any pair ${I}_{i,j},{I}_{k,l} \in {I}$ such that $(i,j)\leq(k,l)$ in lexicographical order. Then the set ${I}'$ is also in $\mathcal{I}_A$, providing ${I}'$ equals ${I}$ except:
\vspace{-1.5\baselineskip}
\begin{align*}
{I}_{i,j}' &:= \oslash \big( {I}_{i,j} \otimes (i,j) + {I}_{k,l} \otimes (k,l) \big) \\
{I}_{k,l}' &:= \cdot 
\end{align*}
\end{description}
\vspace{-0.2\baselineskip}

\noindent This shows we can add any two instances together, in any order, as they are by definition always non-overlapping and thus valid in $A$, and hereby obtain another element of $\mathcal{I}_A$. Eventually this results in just one big instance that is equal to $A$. Note that when we take two elements ${I}_{i,j},{I}_{k,l} \in {I}$ we force $(i,j)\leq(k,l)$, not only to eliminate different routes to the same instance matrix, but also so that the pivot of the new pattern coincides with ${I}_{i,j}$. We can then leave ${I}_{k,l}$ empty.

The construction of $\mathcal{I}_A$ also implicitly defines $\mathcal{H}_A$. While this may seem odd---defining models for instantiations instead of the other way around---note that there is no unambiguous way to find one instantiation for a given model. Instead we find the following definition by applying the inductive construction:
%
\begin{align}
\mathcal{H}_A=\big\{\{\oslash({x}) \ | \ {x} \in {I} \} \ \big | \ {I} \in \mathcal{I}_A \big\}.
\end{align}

\begin{figure}[t]
\begin{tikzpicture}[on grid, grow=right]
\node(R)[l1] {\tiny $\begin{matrix}\overbrace{\begin{bmatrix}0\end{bmatrix}}^{X},\overbrace{\begin{bmatrix}1\end{bmatrix}}^{Y}\\[1em]\underbrace{\begin{bmatrix}X & Y \\Y & X\end{bmatrix}}_{I}\end{matrix}$}
	child {
        node(F)[l3] {\tiny $\begin{bmatrix}0 & 1\end{bmatrix},\begin{bmatrix}V & \cdot \\Y & X\end{bmatrix}$}        
            edge from parent[draw=none] 
            (R) edge (F.west)
    }
    child {
        node(E)[l2] {\tiny $\begin{bmatrix}1 \\ 0\end{bmatrix},\begin{bmatrix}W & V \\ Y & \cdot\end{bmatrix}$}        
            child {
                node(J)[l3]
                    {\tiny $\begin{bmatrix}0 & 1 \\ \cdot & 0\end{bmatrix},\begin{bmatrix}W & \cdot \\ Y & \cdot\end{bmatrix}$}
                edge from parent 
            }
            edge from parent[draw=none] 
            (R) edge (E.west)
    }    
    child {
        node(D)[l2] {\tiny $\begin{bmatrix}\cdot & 1 \\ 1 & \cdot\end{bmatrix},\begin{bmatrix}X & V \\ \cdot & X\end{bmatrix}$}        
            child {
                node(I)[l3]
                    {\tiny $\begin{bmatrix}\cdot & 1 \\ 1 & 0\end{bmatrix},\begin{bmatrix}X & W \\ \cdot & \cdot\end{bmatrix}$}
                edge from parent 
            }
            edge from parent[draw=none] 
            (R) edge (D.west)
    }    
    child {
        node(C)[l2] {\tiny $\begin{bmatrix}0 \\ 1\end{bmatrix},\begin{bmatrix}V & Y \\\cdot & X\end{bmatrix}$}        
            child {
                node(H)[l3]
                    {\tiny $\begin{bmatrix}0 & 1 \\ 1 & \cdot\end{bmatrix},\begin{bmatrix}W & \cdot \\ \cdot & X\end{bmatrix}$
                    }                    
                    	child {
                    		node(K)[l3] {\tiny $\overbrace{\begin{bmatrix}0 & 1 \\ 1 & 0\end{bmatrix}}^{Z},\overbrace{\begin{bmatrix}Z & \cdot \\ \cdot & \cdot\end{bmatrix}}^{I}$}
                    	}
                edge from parent 
            }
            edge from parent[draw=none] 
            (R) edge (C.west)
    }    
    child {
        node(B)[l2] {\tiny $\begin{bmatrix}0 & \cdot \\ \cdot & 0\end{bmatrix},\begin{bmatrix}V & Y \\Y & \cdot\end{bmatrix}$}        
            child {
                node(G)[l3]
                    {\tiny $\overbrace{\begin{bmatrix}0 & \cdot \\ 1 & 0\end{bmatrix}}^{W},\overbrace{\begin{bmatrix}W & Y \\ \cdot & \cdot\end{bmatrix}}^{I}$\vspace{.5cm}}
                edge from parent 
            }
            edge from parent[draw=none] 
            (R) edge (B.west)
    } 
    child {
        node(A)[l2] {\vspace{.5cm}\tiny $\overbrace{\begin{bmatrix}1 &0\end{bmatrix}}^{V},\overbrace{\begin{bmatrix}X & Y \\V & \cdot\end{bmatrix}}^{I}$}        
        	edge from parent[draw=none] 
            (R) edge (A.west)
    };
\draw(F.east)--(H.west);
\draw(F.east)--(J.west);
\draw(B.east)--(J.west);
\draw(C.east)--(G.west);
\draw(D.east)--(H.west);
\draw(E.east)--(I.west);
\draw(A.east)--(G.west);
\draw(A.east)--(I.west);
\draw(G.east)--(K.west);
\draw(I.east)--(K.west);
\draw(J.east)--(K.west);
\end{tikzpicture}
\caption{Model space lattice for a $2\times 2$ Boolean matrix. The V, W, and Z columns show which pattern is added in each step, while $I$ depicts the current instantiation.}
\label{lattice}
\end{figure}
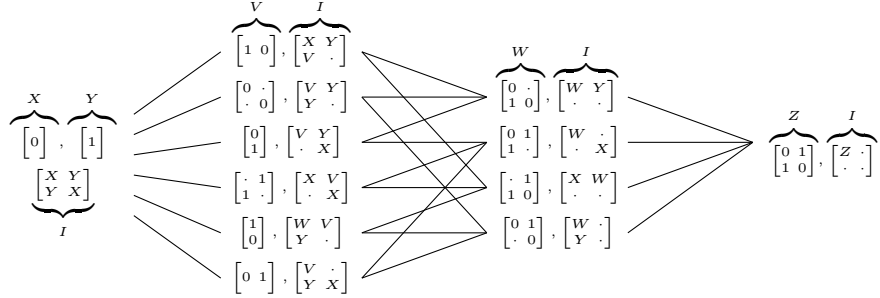

\noindent So for any instantiation ${I}\in \mathcal{I}_A$ there is a corresponding set in $\mathcal{H}_A$ of all patterns that occur in ${I}$. This results in an interesting symbiosis between model and instantiation: increasing the complexity of one decreases that of the other. This construction gives a tightly connected lattice as shown in Figure \ref{lattice}. 

\subsection{Encoding Models and Instances}

From all models in $\mathcal{H}_A$ we want to select the model that describes $A$ best. Two-part MDL \cite{grunwaldmdl} tells us to choose that model that minimises the sum of  $L_1(H_A) + L_2(A|H_A)$, where $L_1$ and $L_2$ are two functions that give the length of the model and the length of `the data given the model', respectively. In this context, the data given the model is given by $I_A$, which represents the accidental information needed to reconstruct the data $A$ from $H_A$.

In order to compute their lengths, we need to decide how to encode $H_A$ and $I$. As this encoding is of great influence on the outcome, we should adhere to the conditions that follow from MDL theory: (1) the model and data must be encoded losslessly; and (2) the encoding should be as concise as possible, i.e., it should be optimal. Note that for the purpose of model selection we only need the length functions; we do not need to actually encode the patterns or data.

\smallskip
\noindent \textbf{Code length functions}.
Although the patterns in $H$ and instantiation matrix $I$ are all matrices, they have different characteristics and thus require different encodings. For example, the size of $I$ is constant and can be ignored, while the sizes of the patterns vary and should be encoded. Hence we construct different length functions\footnote{We calculate code lengths in bits and therefore all logarithms have base 2.} for the different components of $H$ and $I$, as listed in Table \ref{tablelength}. 

When encoding $I$, we observe that it contains each pattern $X\in H$ multiple times, given by the \textbf{usage} of $X$. Using the \textbf{prequential plug-in code} \cite{grunwaldmdl} to encode $I$ enables us to omit encoding these usages separately, which would create unwanted bias. The prequential plug-in code gives us the following length function for $I$. We use $\epsilon = 0.5$ and elaborate on its derivation in the Appendix\footnote{The appendix is available on \url{https://arxiv.org/abs/1911.09587}.}.
\begin{align}
\label{plugin}
	L_{pp}({I}\mid P_{plugin}) &= -\sum^{|H|}_{X_i \in h} \left[ \log \frac{\Gamma(\mathrm{usage}(X_i)+\epsilon)}{\Gamma(\epsilon)}\right] + \log \frac{\Gamma(|{I}| + \epsilon|H|)}{\Gamma(\epsilon|H|)}
\end{align}

\begin{table}[t]
\centering
\vspace{-\baselineskip}
\caption{Code length definitions. Each row specifies the code length given by the first column as the sum of the remaining terms.}
\label{tablelength}
\vspace{3pt}
\begin{tabular*}{\textwidth}{l @{\extracolsep{\fill}}lclcl}
\toprule
 & Matrix  &  Bounds & \# Elements & Positions & Symbols \\ 
\midrule
$L_p(X)$ & Pattern & $\log(MN)$ & \multicolumn{2}{c}{$L_{\mathbb{N}}\binom{M_XN_X}{|X|}$} & $|X| \log(|S|)$\\
$L_1(H)$ & Model & \emph{N/A} & $L_N(|H|)$ & \emph{N/A} & $\sum_{X \in H} L_p(X)$ \\
$L_2({I})$ & Instantiation & \emph{constant} & $\log(MN)$ & \emph{implicit} & $L_{pp}({I})$\\
\bottomrule
\end{tabular*}
\end{table}

Each length function has four terms. First we encode the total size of the matrix. Since we assume $MN$ to be known/constant, we can use this constant to define the uniform distribution $\frac{1}{MN}$, so that $\log{MN}$ encodes an arbitrary index of $A$. Next we encode the number of elements that are non-empty. For patterns this value is encoded together with the third term, namely the positions of the non-empty elements. We use the previously encoded $M_XN_X$ in the binominal function to enumerate the ways we can place the $|X|$ elements onto a grid of $M_XN_X$. This gives us both \emph{how many} non-empties there are as well as \emph{where} they are. Finally the fourth term is the length of the actual symbols that encode the elements of matrix. In case we encode single elements of $A$, we assume that each unique value in $A$ occurs with equal probability; without other prior knowledge, using the uniform distribution has minimax regret and is therefore optimal. For the instance matrix, which encodes symbols to patterns, the prequential code is used as demonstrated before. Note that $L_N$ is the universal prior for the integers \cite{univinteger}, which can be used for arbitrary integers and penalises larger integers.



\section{The Vouw Algorithm}

Pattern mining often yields vast search spaces and geometric pattern mining is no exception. We therefore use a heuristic approach, as is common in MDL-based approaches \cite{krimp,slim,classy}. We devise a greedy algorithm that exploits the inductive definition of the search space as shown by the lattice in Figure~\ref{lattice}. We start with a completely underfit model (leftmost in the lattice), where there is one instance for each matrix element. Next, in each iteration we combine two patterns, resulting in one or more pairs of instances to be merged (i.e., we move one step right in the lattice). In each step we merge the pair of patterns that improves compression most, and we repeat this until no improvement is possible.


\subsection{Finding candidates}

The first step is to find the `best' \textbf{candidate} pair of patterns for merging (Algorithm 1). A candidates is denoted as a tuple $(X,Y,\delta)$, where $X$ and $Y$ are patterns and $\delta$ is the relative offset of $X$ and $Y$ as they occur in the data.
Since we only need to consider pairs of patterns and offsets that actually occur in the instance matrix, we can directly enumerate candidates from the instantiation matrix and never even need to consider the original data.  

The \textbf{support} of a candidate, written $\mathrm{sup}(X,Y,\delta)$, tells how often it is found in the instance matrix. Computing support is not completely trivial, as one candidate occurs multiple times in `mirrored' configurations, such as $(X,Y,\delta)$ and $(Y,X,-\delta)$, which are equivalent but can still be found separately. Furthermore, due to the definition of a pattern, many potential candidates cannot be considered by the simple fact that their elements are not adjacent. 

\smallskip \noindent \textbf{Peripheries}. For each instance $x$ we define its \emph{periphery}: the set of instances which are positioned such that their union with $x$ produces a valid pattern. This set is split into the \emph{anterior-} $\mathrm{ANT}(X)$ and \emph{posterior} $\mathrm{POST}(X)$ peripheries, containing instances that come before and after $x$ in lexicographical order, respectively. This enables us to scan the instance matrix once, in lexicographical order. For each instance $x$, we only consider the instances $\mathrm{POST}(x)$ as candidates, thereby eliminating any (mirrored) duplicates. 

\smallskip \noindent \textbf{Self-overlap}. Self-overlap happens for candidates of the form $(X,X,\delta)$. In this case, too many or too few copies may be counted. Take for example a straight line of five instances of $X$. There are four unique pairs of two $X$'s, but only two can be merged at a time, in three different ways. Therefore, when considering candidates of the form $(X,X,\delta)$, we also compute an \emph{overlap coefficient}. This coefficient $e$ is given by $e = (2N_X+1)\delta_i + \delta_j + N_X$, which essentially transforms $\delta$ into a one-dimensional coordinate space of all possible ways that $X$ could be arranged \emph{after} and \emph{adjacent} to itself. For each instance $x_1$ a vector of bits $V(x)$ is used to remember if we have already encountered a combination $x_1,x_2$ with coefficient $e$, such that we do not count a combination $x_2,x_3$ with an equal $e$. This eliminates the problem of incorrect counting due to self-overlap.

\begin{figure*}[t]
\vspace{-\baselineskip}
\begin{minipage}[t]{.45\textwidth}
	\begin{algorithm}[H]
	\caption{FindCandidates}
	\begin{algorithmic}[1]
	\Require $I$
	\Ensure $C$
	\ForAll{$x \in I$}
		\ForAll{$y \in \mathrm{POST}(x)$}
			\State $X \gets \oslash(x), \ Y \gets \oslash(y)$
			\State $\delta \gets \mathrm{dist}(X,Y)$
			\If{$X = Y$}
				\IfContinue{$V(x)[e] = 1$}
				\State $V(y)[e] \gets 1$
			\EndIf
			\State $C \gets C \ \cup \ (X,Y,\delta)$
			\State $\mathrm{sup}(X,Y,\delta)$ += 1
		\EndFor
	\EndFor
	\end{algorithmic}
	\end{algorithm}%
\end{minipage}
\begin{minipage}[t]{.55\textwidth}
	\begin{algorithm}[H]
	\caption{Vouw}
	\label{alg:vouw}
	\begin{algorithmic}[1]
	\Require $H,\ I$
		\State $C \ \gets$ FindCandidates(I)
		\State $(X,Y,\delta) \in C : \forall_{c \in C} \Delta L((X,Y,\delta)) \leq \Delta L(c)$
	\State $\Delta L_{best} = \Delta L((X,Y,\delta))$
	\If{$\Delta L_{best} > 0 $}
		\State $Z \gets \oslash(X\otimes(0,0) + (Y\otimes\delta))$
		\State $H \gets H \cup \{Z\}$
		\ForAll{$x_i \in I \mid \oslash(x_i)=X$}
			\ForAll{$y \in \mathrm{POST}(x_i) \mid \oslash(y) = Y$}
				\State $x_i \gets Z$,  $y \gets \cdot$
			\EndFor
		\EndFor
	\EndIf
	\State \textbf{repeat until} $\Delta L_{best} \ < \ 0$
	\end{algorithmic}%
	\vspace{-1.75pt}
	\end{algorithm}
\end{minipage}
\end{figure*}

\subsection{Gain computation}

After candidate search we have a set of candidates $C$ and their respective supports. The next step is to select the candidate that gives the best \emph{gain}: the improvement in compression by merging the candidate pair of patterns. For each candidate $c=(X,Y,\delta)$ the gain $\Delta L(A',c)$  is comprised of two parts: (1) the negative gain of adding the union pattern $Z$ to the model $H$, resulting in $H'$, and (2) the gain of replacing all instances $x,y$ with relative offset $\delta$ by $Z$ in $I$, resulting in $I'$. We use length functions $L_1, L_2$ to derive an equation for gain:
\begin{align}
\label{gain}
\begin{split}
	\Delta L(A',c) &= \Big(L_1(H') + L_2(I') \Big) - \Big(L_1(H) + L_2(I) \Big) \\
			    &= L_0(|H|) - L_0(|H|+1) - L_p(Z) + \Big(L_2(I') - L_2(I) \Big)
\end{split}
\end{align}

As we can see, the terms with $L_1$ are simplified to $- L_p(Z)$ and the model's length because $L_1$ is simply a summation of individual pattern lengths. The equation of $L_2$ requires the recomputation of the entire instance matrix' length, which is expensive considering we need to perform it for \emph{every candidate}, \emph{every iteration}. However, we can rework the function $L_{pp}$ in Equation (\ref{plugin}) by observing that we can isolate the logarithms and generalise them into
\begin{align}
	\log_G(a,b) = \log \frac{\Gamma(a+ b\epsilon)}{\Gamma(b\epsilon)} = \log \Gamma(a+ b\epsilon) - \log \Gamma(b\epsilon),
\end{align} 

\noindent which can be used to rework the second part of Equation (\ref{gain}) in such way that the gain equation can be computed in constant time complexity.

\begin{align}
\begin{split}
	L_2(I') - L_2(I) = &\log_G(U(X),1) + \log_G(U(Y),1) \\
			      &- \log_G(U(X)-U(Z),1) - \log_G(U(Y)-U(Z),1) \\
			      &- \log_G(U(Z),1) + \log_G(|I|,|H|) - \log_G(|I'|,|H'|) \\
\end{split}   
\end{align}

\noindent Notice that in some cases the usages of $X$ and $Y$ are equal to that of $Z$, which means additional gain is created by removing $X$ and $Y$ from the model. 

\subsection{Mining a Set of Patterns}

In the second part of the algorithm, listed in Algorithm~\ref{alg:vouw}, we select the candidate $(X,Y,\delta)$ with the largest gain and merge $X$ and $Y$ to form $Z$, as explained in Section~\ref{constructpatterns}. We linearly traverse $I$ to replace all instances $x$ and $y$ with relative offset $\delta$  by instances of $Z$. $(X,Y,\delta)$ was constructed by looking in the posterior periphery of all $x$ to find $Y$ and $\delta$, which means that $Y$ always comes after $X$ in lexicographical order. The pivot of a pattern is the first element in lexicographical order, therefore $\mathrm{pivot}(Z) = \mathrm{pivot}(X)$. This means that we can replace all matching $x$ with an instance of $Z$ and all matching $y$ with $\cdot$. 

\subsection{Improvements}
\label{improvements}


\smallskip
\noindent \textbf{Local search}. 
To improve the efficiency of finding large patterns without sacrificing the underlying idea of the original heuristics, we add an optional local search. Observe that without local search, Vouw generates a large pattern $X$ by adding small elements to an incrementally growing pattern, resulting in a behaviour that requires up to $|X|-1$ steps. To speed this up, we can try to `predict' which elements will be added to $X$ and add them immediately. After selecting candidate $(X,Y,\delta)$ and merging $X$ and $Y$ into $Z$, for all $m$ resulting instances $z_i \in {z_0,\dots,z_{m-1}}$ we try to find pattern $W$ and offset $\delta$ such that
\begin{align}
\label{floodfill}
\forall_{i\in 0\dots m} \exists_w \in \mathrm{ANT}(z_i) \cup \mathrm{POST}(z_i) \ \cdot \ \oslash(w) = W \land dist(z_i, w) = \delta.
\end{align}
\noindent This yields zero or more candidates $(Z,W,\delta)$, which are then treated as any set of candidates: candidates with the highest gain are iteratively merged until no candidates with positive gain exist. This essentially means that we run the baseline algorithm only on the peripheries of all $z_i$, with the condition that the support of the candidates is equal to that of $Z$. 



\smallskip \noindent \textbf{Reusing candidates}.  We can improve performance by reusing the candidate set and slightly changing the search heuristic of the algorithm. The \textbf{Best-*} heuristic selects multiple candidates on each iteration, as opposed to the baseline \textbf{Best-1} heuristic that only selects a single candidate with the highest gain. Best-* selects candidates in descending order of gain until no candidates with positive gain are left. Furthermore we only consider candidates that are all \emph{disjoint}, because when we merge candidate $(X,Y,\delta)$, remaining candidates with $X$ and/or $Y$ have unknown support and therefore unknown gain.




\section{Experiments}

\begin{figure}[t]
\centering
\begin{subfigure}[t]{0.25\textwidth}
\centering
\includegraphics[scale=.9]{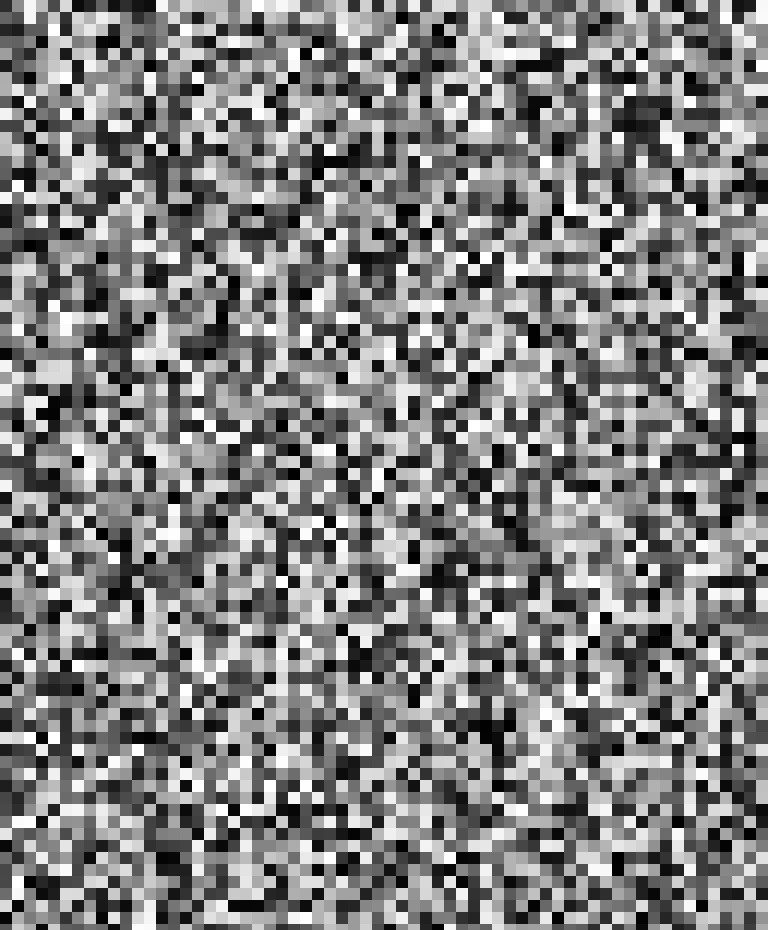}
\caption{Generated matrix}
\label{fig:rila}
\end{subfigure}%
~
\begin{subfigure}[t]{0.25\textwidth}
\centering
\includegraphics[scale=.9]{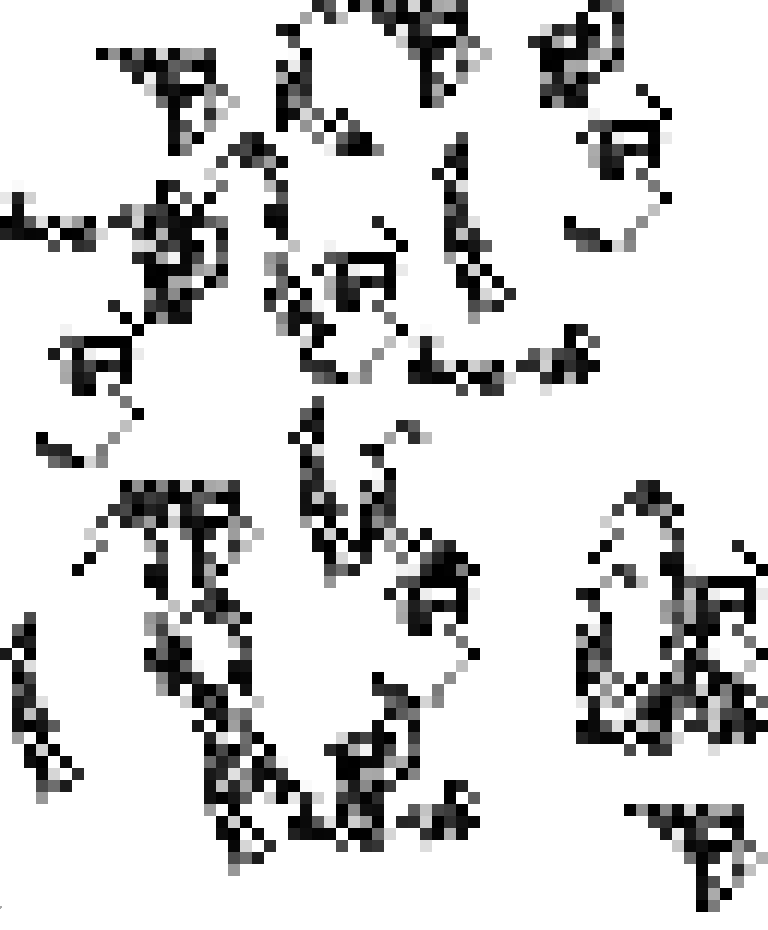}
\caption{Ground truth}
\label{fig:rilb}
\end{subfigure}%
~
\begin{subfigure}[t]{0.25\textwidth}
\centering
\includegraphics[scale=.9]{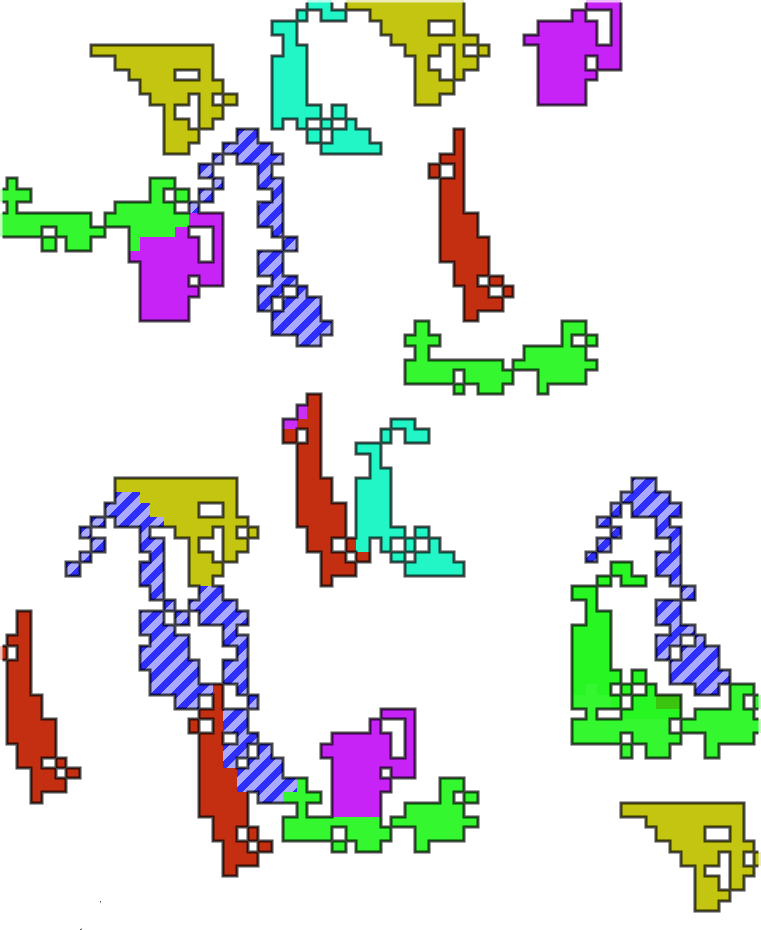}
\caption{Found patterns}
\label{fig:rilc}
\end{subfigure}%
~
\begin{subfigure}[t]{0.25\textwidth}
\centering
\includegraphics[scale=.9]{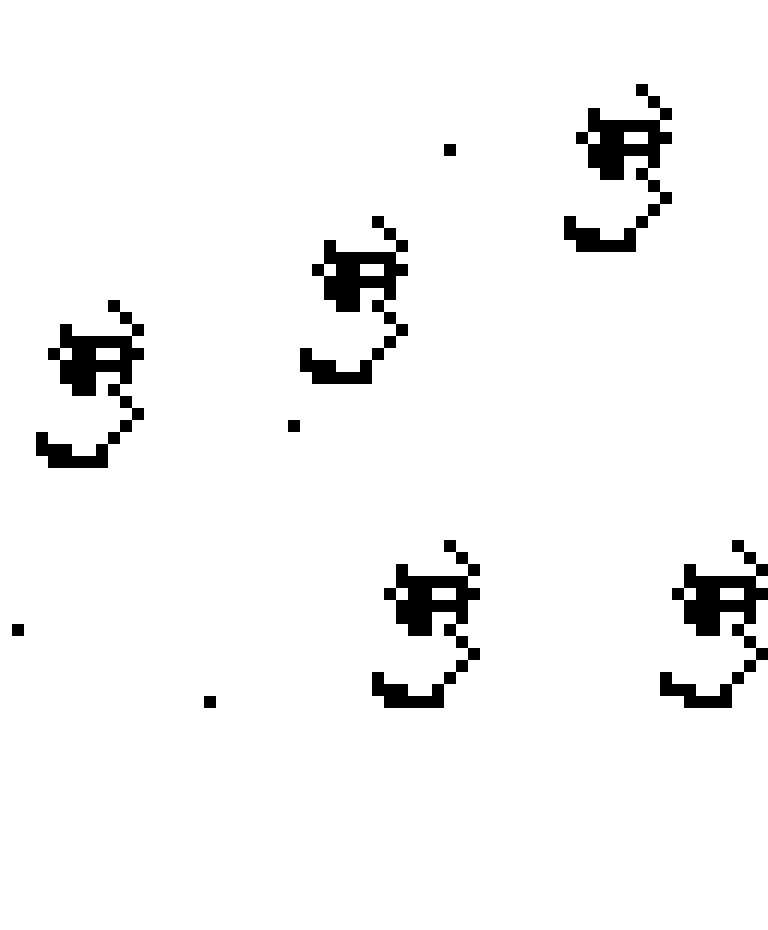}
\caption{Difference}
\label{fig:rild}
\end{subfigure}%
\caption{Synthetic patterns are added to a matrix filled with noise. The difference between the ground truth and the matrix reconstructed by the algorithm is used to compute precision and recall.}
\label{fig:ril}
\end{figure}  

\begin{figure}[t]%
	\begin{subfigure}[t]{0.5\textwidth}
	\includegraphics[scale=1]{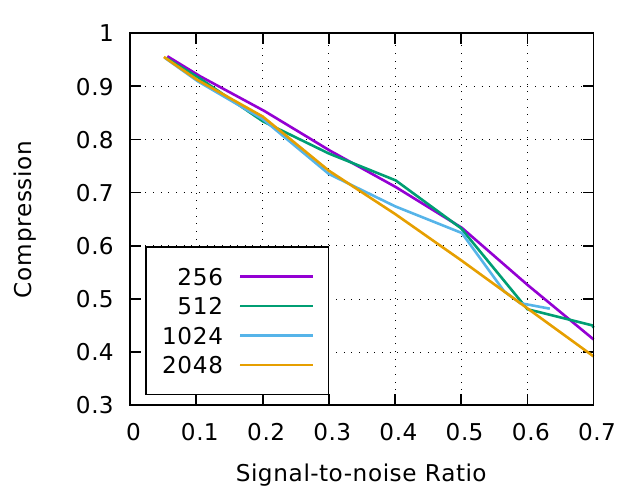}

	\end{subfigure}%
	~
	\begin{subfigure}[t]{0.5\textwidth}
	\includegraphics[scale=1]{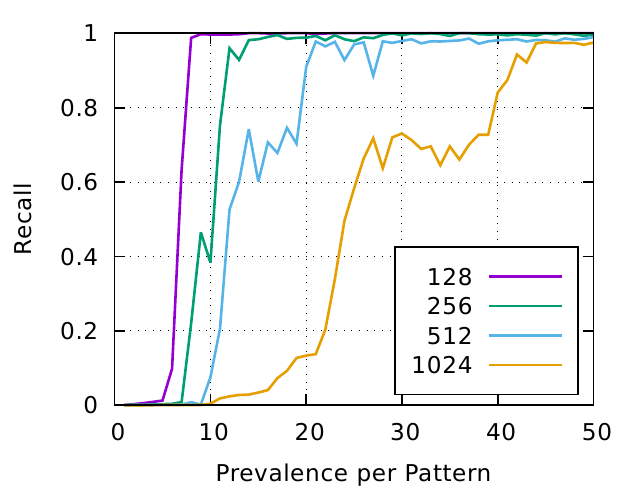}
	\end{subfigure}
	\caption{The influence of SNR in the ground truth (left) and prevalence on recall (right).} 
	\label{fig:plots}
\end{figure}

To asses Vouw's practical performance we primarily use Ril, a synthetic dataset generator developed for this purpose. Ril utilises random walks to populate a matrix with patterns of a given size and prevalence, up to a specified density, while filling the remainder of the matrix with noise. Both the pattern elements and the noise are picked from the same uniform random distribution on the interval $[0,255]$. The \emph{signal-to-noise ratio} (SNR) of the data is defined as the number of pattern elements over the matrix size $MN$. The objective of the experiment is to assess whether Vouw recovers all of the signal (the patterns) and none of the noise. Figure \ref{fig:ril} gives an overview of what the generated data looks like, and how it is mined and evaluated.

\smallskip \noindent \textbf{Implementation.} %
The implementation\footnote{\url{https://github.com/mickymuis/libvouw}} used consists of the Vouw algorithm (written in vanilla C/C++), a GUI, and the synthetic benchmark Ril. Experiments were performed on an Intel Xeon-E2630v3 with 512GB RAM.

\smallskip \noindent \textbf{Evaluation.} %
Completely random data (noise) is unlikely to be compressed. The SNR tells us how much of the data is noise and thus conveniently gives us an upper bound of how much compression could be achieved. We use the ground truth SNR versus the resulting compression ratio as a benchmark to tell us how close we are to finding all the structure in the ground truth. 

In addition, we also compare the ground truth matrix to the obtained model and instantiation. As singleton patterns do not yield any compression over the baseline model, we reconstruct the matrix omitting any singleton patterns. Ignoring the actual values, this gives us a Boolean matrix with `positives' (pattern occurrence=signal) and `negatives' (no pattern=noise). By comparing each element in this matrix with the corresponding element in the ground truth matrix, \emph{precision} and \emph{recall} can be calculated and evaluated.

Figure~\ref{fig:plots} (left) shows the influence of ground truth SNR on compression ratio for different matrix sizes. Compression ratio and SNR are clearly strongly correlated. Figure~\ref{fig:plots} (right) shows that patterns with a low prevalence (i.e., number of planted occurrences) have a lower probability of being `detected' by the algorithm as they are more likely to be accidental/noise. Increasing the matrix size also increases this threshold. In Table \ref{table:optimize} we look at the influence of the two improvements upon the baseline algorithm as described in Section \ref{improvements}. In terms of quality, local search can improve the results quite substantially while Best-* notably \emph{lowers} precision. Both improve speed by an order of magnitude.


\section{Conclusions}

We introduced geometric pattern mining, the problem of finding recurring structures in discrete, geometric matrices, or raster-based data. 
Further, we presented Vouw, a heuristic algorithm for finding sets of geometric patterns that are good descriptions according to the MDL principle. The baseline algorithm is capable of accurately recovering patterns from synthetic data, and the resulting compression ratios are on par with the expectations based on the density of the data. Of the two improvements, especially the local search appears valuable as it improves precision and recall as well as runtime. For the future, we think that extensions to fault-tolerant patterns and clustering have large potential.

\begin{table}[t]
\caption{Performance measurements for the baseline algorithm and its optimizations.}
\label{table:optimize}
\begin{tabular*}{\textwidth}{l @{\extracolsep{\fill}}lccccrrrr}
\toprule
 & & \multicolumn{4}{c}{Precision/Recall} & \multicolumn{4}{c}{Average time}\\
 \cmidrule(l){3-6} \cmidrule(l){7-10} 
 Size & SNR & None & Local & Best-* & Both & None & Local & Best-* & Both \\
\midrule
 256 & .05 & .98/.98 & .99/.99 & .93/.98 & .95/.99 & 29s & 1s & 2s & 1s \\
   & .3 &.99/.8 & .99/.88 & .96/.82 & .99/.89 & 2m 32s & 9s & 5s & 5s \\
 512 & .05 & .98/.97 & .99/.99 & .87/.97 & .93/.98 & 5m 26s & 8s & 20s & 6s \\
  & .3 &.97/.93 & .99/.99 & .94/.91 & .97/.90 & 26m 52s & 2m 32s & 24s & 65s \\
 1024 & .05 & .97/.98 & .99/.99 & .84/.98 & .92/.96 & 21m 34s & 44s & 37s & 34s \\
 & .3 &.98/.98 & .99/.99 & .93/.96 & .98/.97 & 116m 4s & 7m 31s & 1m 49s & 3m 31s \\
\bottomrule
\end{tabular*}
\end{table}



\bibliography{bib}{}
\nocite{*}
\bibliographystyle{plain}

\appendix

\section{Appendix}
\label{appendix_a}
\subsection{Prequential Plugin-Code}

To encode the instance matrix we use the \textbf{prequential plug-in code} \cite{grunwaldmdl}. The prequential plug-in code is defined for sequences of one item at a time and updates the probability of each item as it is encoded, such that the probability need not be known in advance. It has the favorable property of being asymptotically equal to the optimal code for large sequences. Say we want to encode all elements ${I}_i \in {I}$, we define:
\begin{align}
P_{plugin}( y_i = {I}_i \mid y^{i-1} ) = \frac{|\{y \in y^{i-1} \mid y = {I}_i\}| + \epsilon }{\sum_{X \in H}|\{y \in y^{i-1} \mid y = X\}| + \epsilon}
\end{align}
Here $y_i$ is the i-th element to be encoded and $y^{i-1}$ is the sequence of elements encoded so far. We initialize the base case (no element has been sent yet) with a pseudocount $\epsilon$, which gives $P_{plugin}( y_1 = {I} \mid y^{0} ) = \frac{\epsilon}{\epsilon|H|}$. We pick $\epsilon=0.5$ as it is used generally with good results.

Let us adapt this principle to the problem of encoding patterns. The first step here is to determine the probability that each unique element (instance of a pattern) in ${I}$ occurs. 

\label{usage}
\smallskip \noindent $\triangleright$
\emph{Given a set of instances ${I}$, we define $\mathrm{usage}(X) = |\{ {I}_i \in {I} \mid {I}_i = X\}|.$}
\smallskip

From this definition we see that the \textbf{usage} of a pattern is a sum of how often it occurs as an instance. We can use this function to simplify things a little by realizing that we actually know the precise number of instances per pattern on the side of the decoder, but not as the decoder. This information can be used to slightly rephrase Equation \ref{plugin} to be able to encode items in arbitrary order. This produces the length function of the instance matrix ${I}$ as follows\footnote{Here we use the fact that we can interchange sums of logarithms with logarithms of products and that those terms can be moved around freely. Moreover we convert the real-valued product sequences to the Gamma function $\Gamma$, which is the factorial function extended to real and complex numbers such that $\Gamma(n) = (n-1)!$.}:
\begin{align}
\begin{split}
	L_{pp}({I}\mid P_{plugin}) &= \sum^{|{I}|}_{i=1} -\log \frac{|\{y \in y^{i-1} \mid y = {I}_i\}| + \epsilon }{\sum_{X \in H}|\{y \in y^{i-1} \mid y = X\}| + \epsilon}\\
	&= \sum^{|H|}_{X_i \in h} -\log \prod^{\mathrm{usage}(X_i)-1}_{j=0} \frac{j+\epsilon}{\sum^{i-1}_{k=1} U(X_k)+j+\epsilon|H|} \\
	&= -\log \frac{\prod^{X_i\in H} \prod^{\mathrm{usage}(X_i)}_{j=0} j + \epsilon}{\prod^{|{I}|-1}_{j=0} j + \epsilon|H|} \\
	&= -\sum^{|H|}_{X_i \in h} \left[ \log \frac{\Gamma(\mathrm{usage}(X_i)+\epsilon)}{\Gamma(\epsilon)}\right] + \log \frac{\Gamma(|{I}| + \epsilon|H|)}{\Gamma(\epsilon|H|)}
\end{split}
\end{align}


\end{document}